\documentclass{svproc}
\usepackage{amsmath}
\def\R{\mathbb{R}}

\def\1{\mathbf{1}}
\def\0{\mathbf{0}}

\def\optlimits{\nolimits}

\usepackage{times}
\usepackage{epsfig}
\usepackage[utf8]{inputenc} 
\usepackage[french,english]{babel}
\usepackage{float}
\usepackage{amssymb}

\usepackage{url}

\usepackage{xparse}

\makeatletter
\NewDocumentCommand{\raisedminus}{m}{%
  \raisebox{0.2em}{$\m@th#1{-}$}%
}
\NewDocumentCommand{\unaryminus}{}{%
  \mathbin{%
    \mathchoice{%
      \raisedminus\scriptstyle
    }{%
      \raisedminus\scriptstyle
    }{%
      \raisedminus\scriptscriptstyle
    }{%
      \raisedminus\scriptscriptstyle
    }%
  }%
}
\makeatother

\usepackage{graphicx, multicol, footmisc}
\begin{document}
\mainmatter              
\title{A First Derivative Potts Model for Segmentation and Denoising Using ILP}
\titlerunning{Segmentation and Denoising using MILP}  
%
\author{Ruobing Shen\inst{1} \and Gerhard Reinelt\inst{1} \and
Stéphane Canu\inst{2}}
\authorrunning{Ruobing Shen et al.} 
%
\tocauthor{Ruobing Shen, Gerhard Reinelt, Stéphane Canu}
\institute{Institute of Computer Science, Heidelberg University, 69120 Heidelberg, Germany\\
\email{ruobing.shen@informatik.uni-heidelberg.de}
\and
LITIS, Normandie University, INSA Rouen, 76800 Rouen, France}

\maketitle              

\begin{abstract}
Unsupervised image segmentation and denoising are two fundamental tasks in image processing. Usually, graph based models such as multicut are used for segmentation and variational models are employed for denoising. Our approach addresses both problems at the same time.
We propose a novel ILP formulation of the first derivative Potts model with the $\ell_1$ data term, where binary variables are introduced to deal with the $\ell_0$ norm of the regularization term. The ILP is then solved by a standard off-the-shelf MIP solver.
Numerical experiments are compared with the multicut problem.
\end{abstract}
\section{Introduction}
\label{sec:intro}
Segmentation is a fundamental task for extracting semantically meaningful regions from an image.
In this paper we consider the problem of partitioning a given image into an unknown number of segments, i.e., we assume that no prototypical features about the image are available, it is 
a so-called \emph{unsupervised image segmentation problem}.
In a general setting this problem is NP-hard. Exact optimization
models such as the \emph{multicut problem}~\cite{prob,kappesg} are based on \emph{integer linear programming} (ILP) and solved using branch-and-cut methods.

Another aspect of image processing is \emph{denoising}. Main tools for denoising are the variational methods like
the approach with Potts priors which was designed to preserve sharp discontinuities (edges) in images
while removing noises.
Given $n$ signals, denote their intensities $y=(y_1,y_2,\ldots,y_{n})$ (e.g.\ grey scale or color values) and define
$w=(w_1,w_2,\ldots,w_{n})$ as the vector of denoised values.
The classical (discrete) Potts model (named after R. Potts~\cite{potts}) has the form
\begin{equation}
\label{TV}
\min_{w} \;\|w-y\|_k + \lambda\|\nabla^1 w\|_0,
\end{equation}
where the first part measures the $\ell_k$ norm difference between~$w$ and~$y$, and the second part measures the number of oscillations in $w$. Recall that the \emph{discrete first derivative} $\nabla^1 x$ of a vector~$x\in\R^n$ is defined as
the $n-1$ dimensional vector $(x_2-x_1,x_3-x_2,\ldots,x_n-x_{n-1})$ and the $\ell_0$ norm 
of a vector gives its number of nonzero entries.
The scalar $\lambda$ is a parameter for regularization.
Recently, various modifications and improvements have been made for the Potts model, see~\cite{Chan2006} for an overview. 

In general, 
solving the discrete Potts model \eqref{TV} is also NP-hard. In~\cite{Fast} local greedy methods are used to solve it. Recently, \cite{bertsimas2016} uses an ILP formulation to deal with the
$\ell_0$ norm for a similar problem in statistics called the \emph{best subset selection problem}.

Motivated by the above mentioned two models, we are interested in simultaneously segmentation and denoising.  We assume that the input is an image given as grey scale values
(RGB images can be easily transformed) for pixels located on an~$m\times n$ grid.
Let $V=\{p_{1,1},\ldots, p_{m,n}\}$ denote this set of pixels. For representing relations between
neighboring pixels, we define the corresponding grid graph $G=(V,E)$ where $E$ contains
edges between pixels which are horizontally or vertically adjacent. 
A general segmentation is a partition of $V$ into sets $\{V_1, V_2, \ldots, V_k\}$ such that
$\cup_{i=1}^kV_i=V$, and $V_i \cap V_j=\emptyset$, $i\neq j$. 
So in graph-theoretical terms the image segmentation problem corresponds to a graph partitioning problem.

The paper is organized as follows. In Section~\ref{sec:1D} we introduce our \emph{mixed integer programming} (MIP) formulation of problem \eqref{TV}
for the 1D signal case. We then review the multicut problem in Section~\ref{sec:multicut}. Section~\ref{sec:2dproblem} presents our main ILP formulation for 2D images and introduces two types of redundant constraints.
Computational experiments of $3$ instances are presented in Section~\ref{sec:experiment}.
Finally, we conclude and point to future work in Section~\ref{sec:conclusion}.

\section{The First Derivative Potts Model: 1D}
\label{sec:1D}
Given $n$ signals $p=(p_1,\ldots,p_n)$ in some interval $D\subseteq\R$ with intensities $y=(y_1,\ldots,y_n)$. 
We call a function $f$ piecewise constant over $D$ if there is a partition of $D$ into subintervals $D_1, \ldots, D_k$ such that $D=\cup_{i=1}^k D_i$, where $D_i\cap D_j=\emptyset$, and $f$ is constant when restricted to~$D_i$.
Throughout the paper, we assume the input images or signals contain noises. The task of segmentation
and denoising then becomes piecewise constant fitting. The fitting value for signal~$p_i$
is denoted $w_i=f(p_i)$.

\begin{figure}[h]
 \centering
 \includegraphics[width=3.8cm]{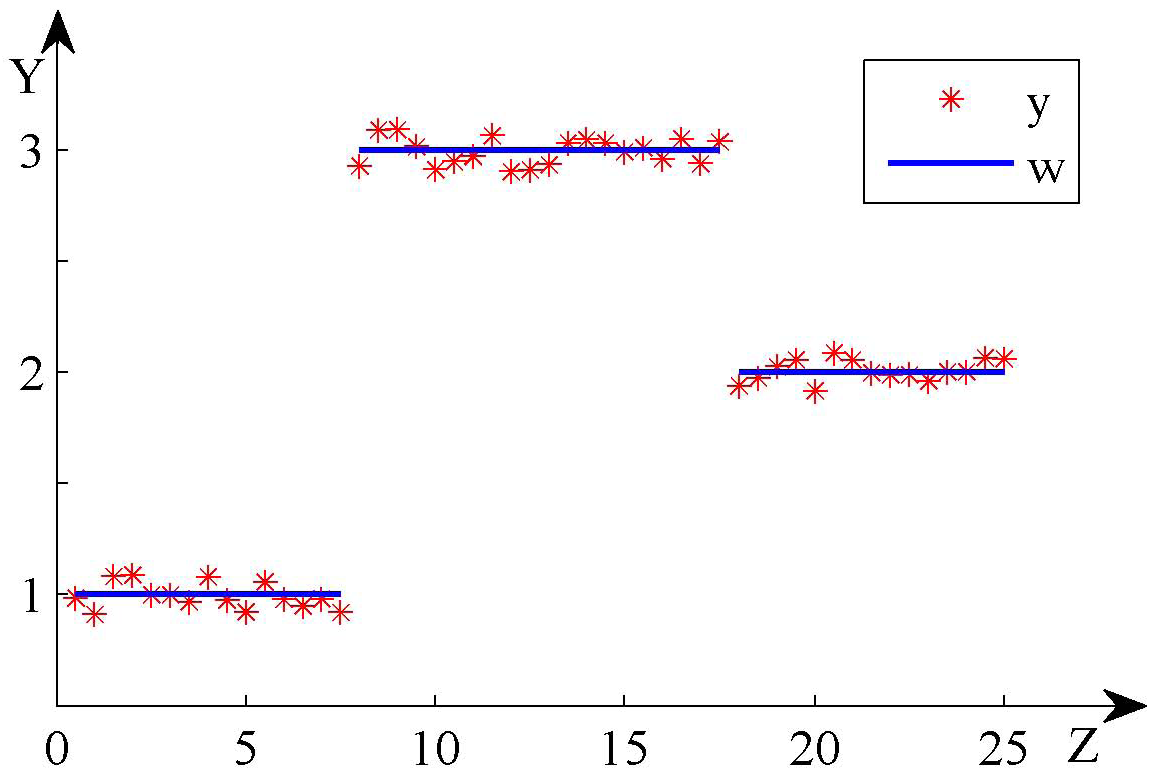}\qquad
 \includegraphics[width=3cm]{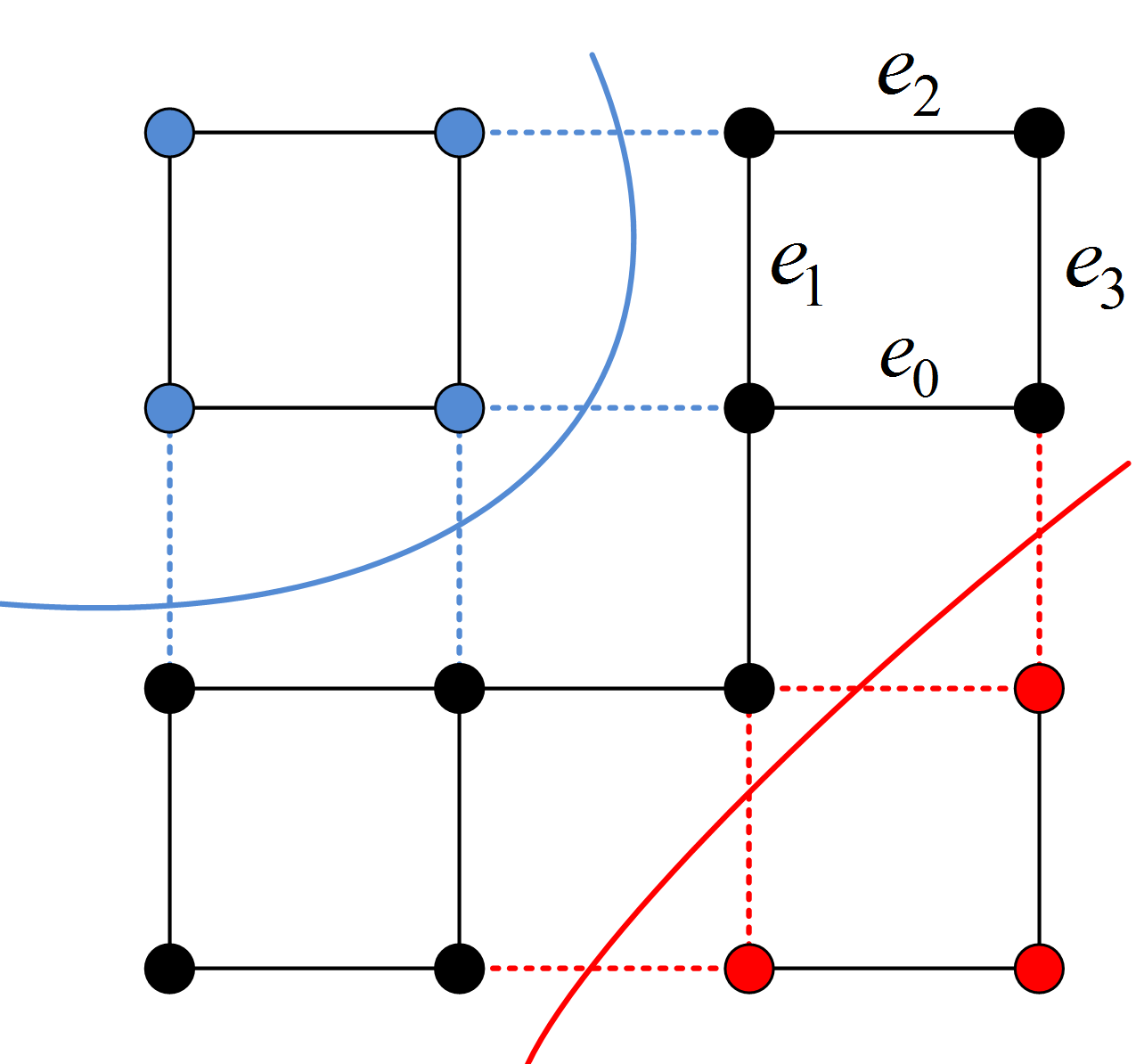}
 \caption{Left: 1D-fitting, 3 segments and 2 active edges. Right: multicut in a $4\times4$-grid.}
 \label{fig:multicut2}
\end{figure}

In 1D, the associated graph $G(V,E)$ is simply a chain, where $V=\{p_i\;| \;i\in[n]\}$
and $E=\{e_{i}= (p_i,p_{i+1})\;|\; i\in [n-1]\}$. Here, $[n]$ denotes the discrete set $\{1,2,\ldots, n\}$.
We propose to formulate~problem \eqref{TV} as an MIP by introducing $n-1$ binary variables~$x_{e_i}$, where  $x_{e_i}=1$ if and only if the end nodes of~$e_i$ are in different segments. If so, the edge
is called \emph{active}, otherwise it is \emph{dormant}. 
Since  $w$ is restricted to be constant within the same segment, it follows that $w_{i+1}-w_i\ne 0$ if and only
if $p_i$ and $p_{i+1}$ are on the boundary (i.e., $x_{e_{i}} = 1$). Thus the signals between two active edges define one segment and the
number of segments is $\sum\optlimits_{i=1}^{n-1} x_{e_i}+1 $. See the left part of Fig.~\ref{fig:multicut2}
for an example, where there are two active edges and three segments.

An MIP formulation for \eqref{TV} is
\begin{alignat}{2}
	\min \;\; \sum\optlimits_{i=1}^n|w_i-&y_i |+\lambda \sum\optlimits_{i=1}^{n-1} x_{e_i} \label{eq:Pott_1D2}\\ 
	|w_{i+1}-w_i| &\leq M x_{e_i}, \quad && i\in[n-1],\tag{\ref{eq:Pott_1D2}a}\label{bigmcons}\\
	w_n &\in \R,\;\; &&i\in[n],\nonumber\tag{\ref{eq:Pott_1D2}b}\\
	x_{e_i} &\in \{0,1\},\;\;&&i\in[n-1]\nonumber\tag{\ref{eq:Pott_1D2}c},
\end{alignat}
where $\lambda$ is the penalty parameter for the number of segments to prevent over-fitting, and $M$ is usually called the ''big M'' constant in MIP to ensure that the constraints~\eqref{bigmcons} are always valid. It enforces that the pixels corresponding to the end nodes of a dormant edge ($x_e=0$)
have the same fitting value. 
Note that we use the $\ell_1$ norm because it can be easily
modeled with linear constraints. Namely, constraint~\eqref{bigmcons} is replaced by the two constraints
$w_{i+1}- w_{i} \leq Mx_{e_i}$ and $-w_{i+1}+ w_{i} \leq Mx_{e_i}$, and the term
$|w_i-y_{i}|$ is replaced by $\varepsilon_i^+ +\varepsilon_i^-$ where $w_i - y_i = \varepsilon^+_i - \varepsilon^-_i$
and $\varepsilon_i^+, \varepsilon_i^- \geq 0.$ Moreover, it is more robust to noise than $\ell_2$.

The solution of~\eqref{eq:Pott_1D2} gives the fitting value~$w_i$ for the signal~$p_i$ and the boundaries of
two segments are given by the active edges ($x_e=1$). 
From now on, for simplicity,
we will just specify models in form~\eqref{eq:Pott_1D2}.

\section{The Multicut Problem}
\label{sec:multicut}
The multicut problem~\cite{prob}
formulates the graph partitioning problem as an \emph{edge labeling} problem. 
For a partition ${\cal V}=\{V_1, V_2, \ldots, V_k\}$ of $V$, the edge set 
$\delta(V_1, V_2, \ldots, V_k) = \{uv \in E \mid \exists i \ne j \text{ with } u\in V_i \text{ and } 
v \in V_j\}$
is called the \emph{multicut} induced by $\cal V$. 
We introduce binary edge variables $x_e$  and represent the multicut by a set of active edges.

With the edge weight $c: E\rightarrow \R$ representing the absolute differences between two pixels' intensities,
the multicut problem~\cite{kappesg} can be formulated as the following ILP
\begin{align}
\min \;\; \sum\optlimits_{e\in E}& \unaryminus c_ex_e + \sum\optlimits_{e\in E}\lambda x_e\label{multicut}\\
\sum\optlimits_{e\in C\setminus\{e'\}}x_e&\geq x_{e'}, \;\;  \forall \;\text{cycles $C\subseteq E$, $e'\in C$},\tag{\ref{multicut}a} \label{multicut_a}\\
x_e &\in \{0,1\},\;\; \forall e\in E, \tag{\ref{multicut}b}
\end{align}
Constraints~\eqref{multicut_a} are called the \emph{multicut constraints} and they enforce the consecutiveness of the active edges, and in turn the connectedness of each segment. 
Thus each maximal
set of vertices induced only by dormant edges corresponds to a segment. 
The right part of Fig.~\ref{fig:multicut2}
shows a partition of a $4\times 4$-grid graph into 3~segments where the dashed active edges form the multicut.

Problem~\eqref{multicut} is NP-hard in general and while the number of inequalities~\eqref{multicut_a} can be exponentially large, violated constraints can be found efficiently using shortest path algorithms. They are added iteratively until the solution is feasible~\cite{prob,kappesg}.

\section{The First Derivative Potts Model in 2D}
\label{sec:2dproblem}


The main formulation is modeled as a discrete first derivative Potts model
and is obtained by formulating \eqref{eq:Pott_1D2} per row and column. We also talk about adding redundant constraints to speed up computation in Sec.~\ref{redun}.


\subsection{Main formulation}
Given a 2D image, following notation from Sec.~\ref{sec:1D}, we further divide $E=E^r\cup E^c$ into its
row (horizontal) edge set $E^r$ and column (vertical) edge set~$E^c$.
Denote $e_{i,j}^r\in E^r$ as the row edge $(p_{i,j},p_{i,j+1})$ and $e_{i,j}^c\in E^c$ as the column edge $(p_{i,j},p_{i+1,j})$. Our main formulation is

\begin{align}
%
\text{min}\;\; \sum\optlimits_{i=1}^m \sum\optlimits_{j=1}^n | w_{i,j} - y_{i,j} |& +\lambda\sum\optlimits_{e\in E} x_e &\label{model2}\\
	|w_{i,j+1}-w_{i,j}| &\leq Mx_{e^r_{ij}}, \;\; i \in [m],\; j\in[n-1],\tag{\ref{model2}a}\\
	|w_{i+1,j}-w_{i,j}| &\leq M x_{e^c_{ij}}, \;\; j\in[n],\; i\in [m-1],\tag{\ref{model2}b}\\
	w_{i,j} &\in \R,   \;\; i \in [m],\; j\in[n],\tag{\ref{model2}c}\\
	x_e  &\in \{0,1\}, \;\;e\in E \tag{\ref{model2}d}.
\end{align}

\subsection{Redundant constraints}
\label{redun}
It is common practice to add redundant constraints to an MIP for computational efficiency. A constraint is redundant if it is not necessarily needed for a formulation to be valid. However, they may be useful because they forbid some fractional solutions during the branch-and-bound approach, where the MIP solver iteratively solves the linear programming (LP) relaxation, or because they impose a structure that help shrink the search space.

It is well known that if a cycle $C\in G$ is chordless,
then the corresponding constraint~\eqref{multicut_a} is facet-defining for the multicut 
polytope~\cite{kappesg}, thus providing the tightest LP relaxation. Inspired by the multicut problem~\eqref{multicut}, in a grid graph, although the number of such constraints is still exponential, it may be advantageous to add the following 4-edge chordless cycle constraints

\begin{equation}
\sum\optlimits_{e\in C\setminus\{e'\}}x_e\geq x_{e'}, \;\;  \forall \;\text{cycles $C\subseteq E$, $|C|=4$, $e'\in C$} \label{multicut4}
\end{equation}
to \eqref{model2}.
Meanwhile, if the user has some prior knowledge or good guesses on the number of active edges, it might be beneficial to add the following cardinality constraints

\begin{equation}
\sum\optlimits x_e\leq k \label{regu}
\end{equation}
per row and column in a given image.

We show detailed experiments in Sec. \ref{sec:experiment} on how adding the above two types of constraints affects the computation.

\begin{figure}[t]
  \center
 \includegraphics[width=11.9cm]{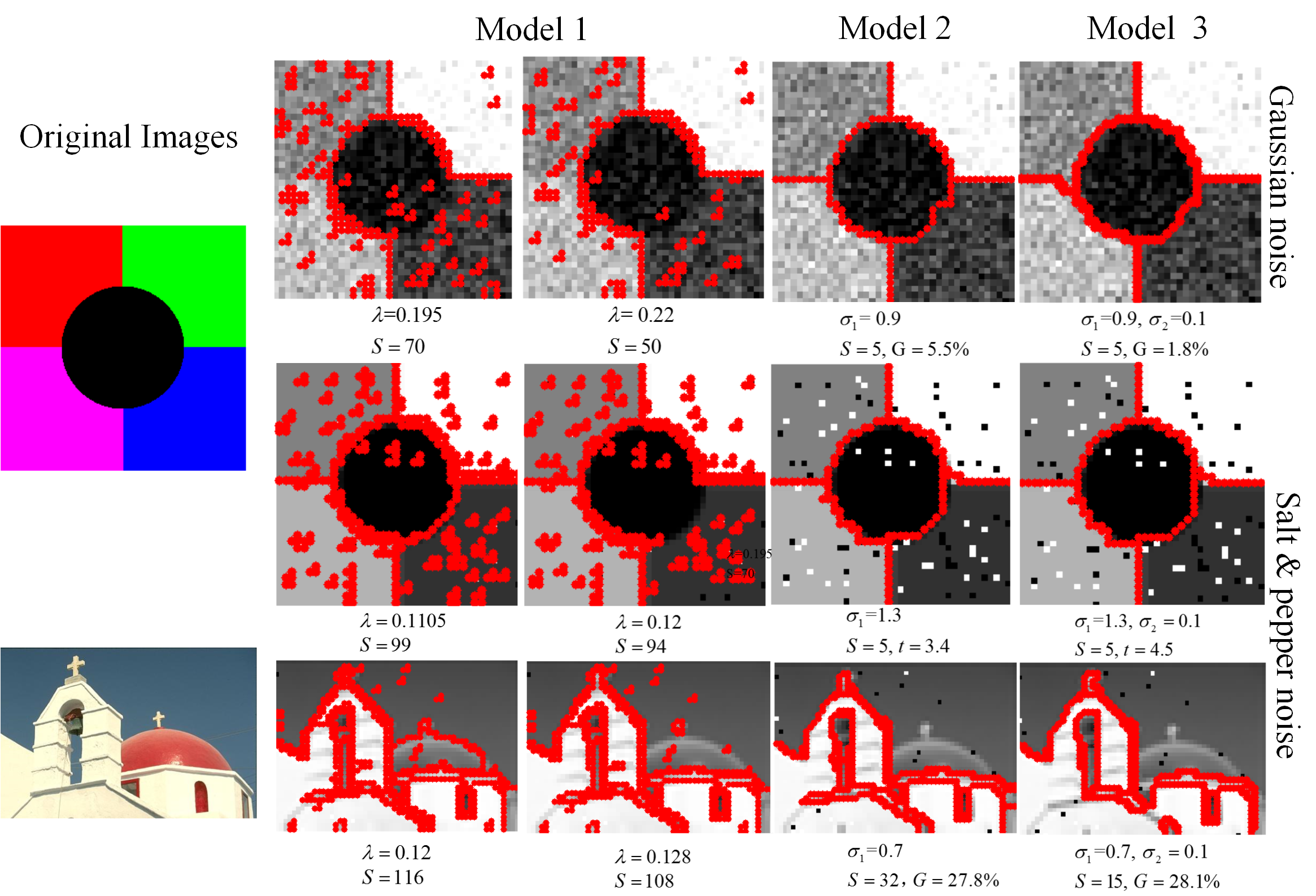}
 \caption{Segmentation results of three models. S: number of segments. t: running time. G: optimality gap when it hits the time limit of 100 sec. }
 \label{fig:ex}
\end{figure}

\section{Computational Experiments}
\label{sec:experiment}
Computational tests are performed using Cplex $12.6.1$, a standard MIP solver, on a Intel i5-4570 quad-core desktop with 16GB RAM. We compare three different models where Model~$1$ is the multicut problem~\eqref{multicut}, Model $2$ refers to our main formulation~\eqref{model2}, with and without the 4-edge cycle constraints~\eqref{multicut4}, and Model $3$ is the main formulation~\eqref{model2} with both~\eqref{multicut4} and~\eqref{regu}.
We take two images from~\cite{martin}, and resize them to $40\times 40$ and $41\times 58$. We add Gaussian and salt and pepper noise, and set a time limit of 100 sec.

\emph{Parameter setting}. We first compute the average intensity of each $4\times 4$ pixels block in the image, and then calculate the absolute difference of its maximum and minimum value, denoted~$Y^{*}$. So~$Y^{*}$
somehow represents the global contrast of the image. 
We set the constant $M$ to ~$Y^{*}$, and $\lambda$ to~$\frac{1}{4}\sigma_1Y^*$, where $\sigma_1$ is a user defined parameter. When there exists an extreme outlier, model \eqref{model2} tends not to treat the single outlier as a separate segment, since doing so would incur a penalty of $4\lambda$.
Denote $ Y^r_i=(y_{i,1}, \ldots, y_{i,n})$, the constant
$k^r_i$ in~\eqref{regu} of row $i$ is set to the number of elements in $\nabla^1 Y^r_i$ that are greater than $\sigma_2 Y^{*}$, where $0<\sigma_2<1$ is some suitably chosen parameter. Constant $k^c_j$ is computed similarly for each column $j$.

Fig.~\ref{fig:ex} shows the input images, detailed setting of the parameters, and the segmentation results for the $3$ models. 

\emph{With and without~\eqref{multicut4}}. We first report that Model $2$ with~\eqref{multicut4} saves $0.9$ sec in the second instance, and narrows $23.5\%$ of the Cplex optimality gap on average, compared to without~\eqref{multicut4}. In later comparison, we denote Model $2$ as the formulation~\eqref{model2} with ~\eqref{multicut4}.

\emph{Running time and optimality gap}. Model $1$ is very fast to solve, takes less than $0.1$ sec in all three instances. Model $2$ and $3$ take $2.4$ and $4.5$ sec in the second instance and hit the time limit in the other two. The optimality gap for Model $2$ and $3$ on the first and the third instance, are $5.5\%$, $27.8\%$, $1.8\%$ and $28.1\%$ respectively.

\emph{Model $2$ versus $3$}. We keep $\sigma_1$ the same when comparing the effects of adding~\eqref{regu}. There is no clear advantage of adding the cardinality constraints~\eqref{regu}, since for example, it enlarges the optimality gap in the third instance while the solution is visually better.

As we can see from Fig.~\ref{fig:ex}, Model $1$ is sensitive to noise and the parameter~$\lambda$. As a result, it is over partitioned, and is hard to control the desired number of segments.
On the other hand, although requiring more computational time, Model $2$ and $3$ are robust to noise, less sensitive to parameters, and give better segmentation results. In addition, we found it beneficial to add the 4-edge cycle constraints~\eqref{multicut4}, while there is no clear conclusion on whether to add the cardinality constraints~\eqref{regu} to~\eqref{model2}.

\section{Conclusions and Future Work}
\label{sec:conclusion}
We present an ILP formulation of a discrete first derivative Potts model with $\ell_1$ data term for simultaneously segmenting and denoising. The model is quite general, firstly, it can use any heuristic method like~\cite{Fast} as an initial solution and provide a guarantee (lower bound) by solving an LP. Secondly, it could improve the initial solution by finding a better solution within the branch-and-bound framework using any MIP solver. 

Decomposition algorithms such as superpixel lattice algorithms~\cite{latticecut} could be used as preprocessing towards larger images. We will also explore the possibilities of applying our model to 3D images. Finally, since the underlying problem is piecewise constant fitting, applications beyond the scope of computer vision are also of interest.
%
%
%
\bibliographystyle{unsrt}
\bibliography{research}

\end{document}